\documentclass[acmtog, authorversion]{acmart}

\usepackage{booktabs} 

\setcitestyle{square}

\usepackage[ruled]{algorithm2e} 

\SetAlFnt{\small}
\SetAlCapFnt{\small}
\SetAlCapNameFnt{\small}
\SetAlCapHSkip{0pt}
\IncMargin{-\parindent}

\acmJournal{TOG}
\acmVolume{0}
\acmNumber{0}
\acmArticle{0}
\acmYear{2018}
\acmMonth{0}

\setcopyright{usgovmixed}

\acmDOI{0000001.0000001_2}

\received{March 2018}
\received{June 2018}
\received[final version]{June 2018}
\received[accepted]{July 2018}

\hypersetup{draft}
\begin{document}
\title{Scalable closed-form trajectories for periodic and non-periodic human-like walking}

\author{Salman Faraji}
\author{Auke J. Ijspeert}
\affiliation{%
\institution{EPFL}
\country{Switzerland}}
\email{salman.faraji@epfl.ch}

\renewcommand\shortauthors{Faraji, S. et al.}

\begin{abstract}
We present a new framework to generate human-like lower-limb trajectories in periodic and non-periodic walking conditions. In our method, walking dynamics is encoded in 3LP, a linear simplified model composed of three pendulums to model falling, swing and torso balancing dynamics. To stabilize the motion, we use an optimal time-projecting controller which suggests new footstep locations. On top of gait generation and stabilization in the simplified space, we introduce a kinematic conversion method that synthesizes more human-like trajectories by combining geometric variables of the 3LP model adaptively. Without any tuning, numerical optimization or off-line data, our walking gaits are scalable with respect to body properties and gait parameters. We can change various parameters such as body mass and height, walking direction, speed, frequency, double support time, torso style, ground clearance and terrain inclination. We can also simulate the effect of constant external dragging forces or momentary perturbations. The proposed framework offers closed-form solutions in all the three stages which enable simulation speeds orders of magnitude faster than real time. This can be used for video games and animations on portable electronic devices with a limited power. It also gives insights for generation of more human-like walking gaits with humanoid robots.
\end{abstract}

\maketitle

\section{Introduction}
The musculoskeletal system of human has multiple degrees of freedom and many muscles used to produce a wide range of activities. In particular, human walking features many complex motions in the lower-limbs produced by gravity and muscle forces. This complex system can be simplified to reproduce walking behaviors in simulation environments, depending on the level of details needed. The limbs can be simulated with multi-segment rigid bodies while rotary actuators in the joints can play the role of muscles to some extent. Such huge simplifications are probably enough to produce very realistic locomotion behaviors, however, a powerful controller is needed to stabilize the gait. A unified framework is hard to achieve given different anatomical properties, gait parameters, styles of motion and environment conditions. Besides, a plausible controller in this framework should easily handle transition conditions as well as capturing disturbances to simulate interactions with the environment. Many successful controllers are proposed in \cite{yin2007simbicon, tsai2010real, coros2010generalized, mordatch2010robust} for example which use motion-capture data, simplified models or dynamic equations to achieve amazing walking behaviors.

In this work, we mainly focus on simulating essential principles of walking in the lower-limbs. We propose a method that combines trajectories of 3LP, a simple walking model developed earlier \cite{faraji20173lp}, with additional features that result in a human-like gait. We use a previously developed controller called time-projection \cite{faraji20173lp2} to stabilize the gait and perform transitions. Recorded human data is also used to validate trajectories quantitatively. Thanks to linearity of the 3LP model and simplicity of the controller, we offer closed-form solutions for all lower-limb trajectories of human walking in a wide range of parameters. Our method, therefore, captures the main principles of walking with microsecond calculations. It can simulate walking behaviors multiple orders of magnitude faster than physics-based simulators which work at best in real time. This could be used for video games and animations especially on portable electronic devices with limited computational capabilities.

Animating walking behaviors is primarily inspired by biomechanics studies which quantify human gait properties and explain the mechanics behind. There are multiple techniques used to reproduce the behavior, ranging from pure interpolation of human data to detailed physics-based simulations that implement low-level control rules to mimic human behavior. This section reviews different walking animation methods proposed in the literature to highlight some key inspiring ideas. We classify the existing literature into three main categories: interpolation of recorded data, artificial trajectory synthesis, and physics-based simulation. We are more interested in a trade-off between genericity of the method and computation times. In this regard, each category provides certain advantages and limitations discussed as follows. 

\subsection{Interpolation of Recorded Data}
Using a motion-capture system and markers placed on different body parts, one can capture human locomotion trajectories in different conditions. Choi et al., for example, used a motion library to plan bipedal locomotion in un-structured environments \cite{choi2003planning}. They planned probabilistic road-maps which determined a sequence of character configurations based on foothold locations. Similarly, Lee et al. used a human motion database to control animated avatars \cite{lee2002interactive}. A Markov process was used in this method to plan motion phases while blending transition rules were applied to produce smooth motions based on a relatively large database of recorded data in non-periodic walking conditions. Similarly, a single set of periodic walking trajectories were scaled in \cite{okada2013motion} and applied to a robot after fixing dynamical consistency. 

Assuming a fast data query from the database, interpolations could be done relatively fast in terms of computations. However, if dynamical consistencies are considered like \cite{okada2013motion}, a small modification of trajectories is needed. Such modification might not be visually notable, but it might require expensive iterative computations. Besides, the quality of animations might heavily depend on the quality of recorded data. The method in \cite{lee2002interactive} for example requires enough data-points to produce transient conditions. It is also sensitive to glitches and slippages. In a broader perspective, generalizing interpolations to characters with different anatomical properties than the dataset is feasible but challenging \cite{hodgins1997adapting}. Covering a wide range of locomotion parameters such as speed, frequency, and style is also hard with a sparse database \cite{hodgins1997adapting}. Besides, a realistic simulation of interactions with the environment seems impossible in this approach due to the absence of physics in the interpolations.

\subsection{Artificial Trajectory Synthesis}

A similar approach is to consider parametric trajectories instead of interpolating a motion database. Given the wide range of kinematic data reported in biomechanics studies, one can design empirical trajectories with appropriate phase and adjustable amplitudes for certain variables in the system. The approach proposed in \cite{boulic1990global} for example considers sinusoidal variations for pelvis translations and rotations. It also considers similar trajectories for the hip, knee, ankle, thorax, shoulder and the elbow joints. A predefined phase and adjustable amplitude (as a function of velocity) for each of these trajectories can produce walking motions in a wide range of speeds. The relation of step frequency and speed is also taken from human data while an inverse kinematics algorithm slightly modifies trajectories to ensure contact constraints. The parametrization process can be done over key-frames too. The approach proposed in \cite{li2000simulating} for example interpolates between different postures defined for certain gait events to obtain continuous trajectories. By changing the key-frames only, plausible walking trajectories can be obtained for different inclined terrains and walking speeds. The speed-frequency relation, as well as the double-support time ratio, is taken from human data in \cite{li2000simulating}. 

Parametric trajectories have a wide range of applications in robotics. Handharu et al. used parametric trajectories for the foot and the knee to produce walking motions with toe and heel joints \cite{handharu2008gait}. Similarly, Ogura et al. used parametric trajectories for the waist roll, foot positions, and the knee joints to produce stretched-knee walking patterns \cite{ogura2003stretch}. The approach proposed in \cite{ogura2006human} used similar prescribed knee trajectories and optimized other parametric trajectories to find a ZMP-stabilized gait, validated on the WABIAN-2 robot. ZMP refers to Zero Moment Point around which contact reaction forces produce no moment in the horizontal direction. Manually designed Center of Mass (CoM) height trajectories together with a robust inverse kinematics method could later produce stretched-knee gaits on a refined version of this robot \cite{kryczka2011stretched} as well. Although dynamic consistencies were resolved in all these methods \cite{handharu2008gait, ogura2003stretch, ogura2006human, kryczka2011stretched} by proper regulation of ZMP trajectories \cite{zmp_full_review2}, no online control method was proposed. Heerden, on the other hand, used sinusoidal reference trajectories for the pelvis and optimized jerks in a Model Predictive Control setup to plan CoM trajectories with variable heights \cite{van2015planning}. This framework offered reactive stepping and online control, though costly in terms of computations, and missing the knee and ankle joints. The method proposed in \cite{griffin2017straight} offers a faster online control based on the capturability control framework \cite{capturability}, though uses manually tuned set-points for the knee joint in different phases of motion to achieve stretched-knee walking. 

Parametric trajectories are easy to design and tune for specific gaits, but hard to generalize for a broader range of walking conditions, e.g., at different speeds, frequencies, inclinations, walking styles, foot lifts, character sizes, etc. So far, we have only discussed trajectory generation methods (interpolation and synthesis) that impose kinematics. An alternative approach would be to compromise the animation speed and simulate system dynamics directly by time-integration to let the kinematics emerge automatically, i.e. taking dynamics into account. 

\begin{figure*}[]
    \centering
    \includegraphics[trim = 0mm 0mm 0mm 0mm, clip, width=1\textwidth]{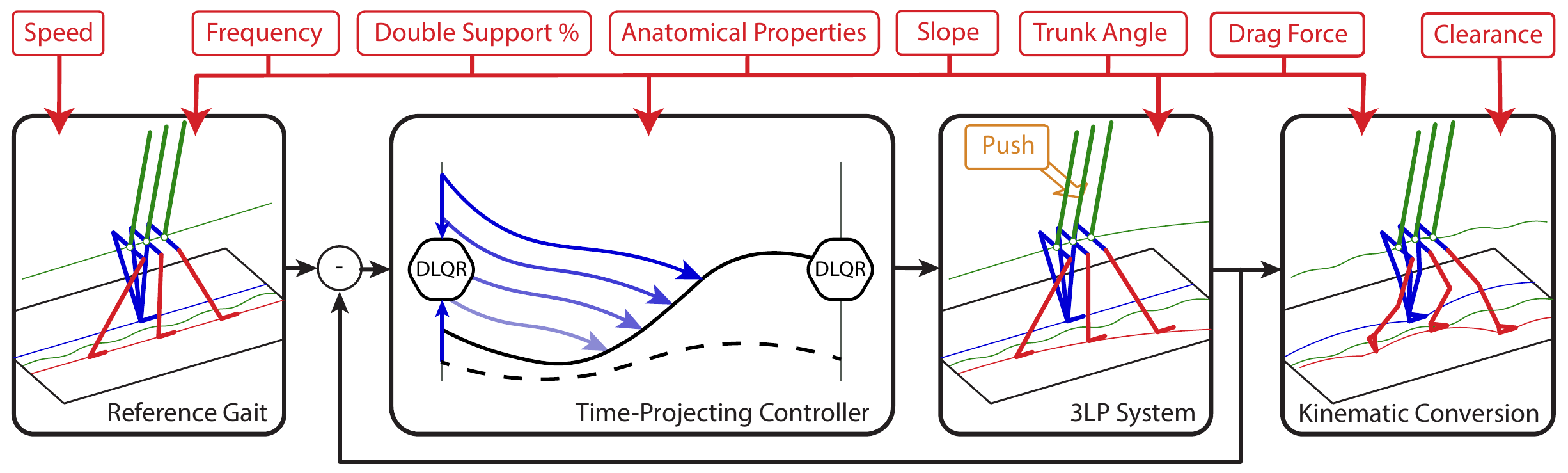} 
    \caption{An overview of our proposed simulation architecture. Given certain walking parameters shown in red, a reference 3LP gait is calculated based on which the time-projecting controller suggests footstep adjustments. The resulting trajectories of the 3LP system are then converted to a more human-like posture. } 
    \label{fig::architecture}
\end{figure*}

\subsection{Physics-Based Simulation}
Simulating the full dynamical model simply translates to a complicated control problem. The walking system has multiple degrees of freedom while the type of task is hybrid, involving a change of mechanical model in each phase of motion (i.e., left, right or double support). Very similar to the few interpolation approaches mentioned earlier, one can use recorded human data or parametric trajectories to produce animations via position control. However, such open-loop controller might only be stable over a small set of states. To tackle this problem, Yin et al. simplified walking dynamics with a single mass and improved stability by suggesting footstep adjustments \cite{yin2007simbicon}. In their simulation framework called SIMBICON, they obtained desired joint angles by applying a Fourier transformation on the recorded human data and taking only essential harmonics. While replicating the recorded trajectories, they used a simple proportional controller in the stance hip to regulate the trunk orientation. A control rule was also introduced to adjust the next footstep location as a linear function of CoM relative position and velocity (with respect to the stance foot) by tunable gains. Using task-specific parameter tunings, SIMBICON was able to achieve realistic walking gaits with different torso styles, leg lifts, motion directions, and push recovery properties. The method proposed by \cite{tsai2010real} was also very similar to SIMBICON and relied on motion-capture data. However, Tsai et al. used an inverted pendulum model to adjust the footstep locations instead of the original tunable gains used on the relative CoM position and velocity in SIMBICON.

The free parameters of the SIMBICON framework together with initial conditions were later optimized by \cite{wang2009optimizing} using mechanical power terms in the objective function to produce more natural walking gaits. The optimized framework removed dependency on the motion-capture data and handled inclined walking as well. However, the optimization procedure had to be repeated for characters with different body shapes. Favoring generalization of the controller, Coros et al. also removed any dependency on the motion-capture data and only used few parametric spline trajectories to allow for human-like knee and ankle trajectories \cite{coros2010generalized}. They used the inverted pendulum model as a core motion generator with parametric swing leg motions. The resulting trajectories were converted to joint-space via inverse kinematics. In the low-level control, he used small gains to track the desired joint angles in addition to gravity compensation and Center of Pressure (CoP) modulation for better compliance and stability. Assuming a decoupling between horizontal and vertical dynamics, Mordatch et al. used a Linear Inverted Pendulum (LIP) and a Spring-Loaded Inverted Pendulum (SLIP) to describe motions in these directions respectively \cite{mordatch2010robust}. They formulated a robust nonlinear Model Predictive Control (MPC) problem to plan the motion, although reaching a reactive online control slowed down their simulations considerably. The complete inverse dynamics formulation used by \cite{mordatch2010robust} could, however, unify the gravity compensation and CoP modulation rules of \cite{coros2010generalized} and produced natural upper-body motions. Apart from simplifying the dimensionality problem, inverse dynamics can provide compliance \cite{faraji2015practical}, realize imprecise Cartesian plans \cite{you2016straight} and allow for multi-character interactions \cite{vaillant2017multi}.

A sub-category of physics-based simulation methods aims at finding task-specific controllers, but not using simplified models. A network of neurons stimulating virtual human-like muscles (neuro-muscular model) can be optimized for example to produce human-like walking gaits at different speeds \cite{geyer2010muscle}. Likewise, a character-specific optimization of control gains and set-point angles for a musculoskeletal model combined with the SIMBICON stabilization rules can also produce realistic walking behaviors \cite{geijtenbeek2013flexible}. Individual controllers can be composed together to cover a wider range of tasks using support vector machines \cite{faloutsos2001composable} or interpolation of control laws \cite{laszlo1996limit}. Off-line optimizations can also achieve more versatility by using exteroceptive sources of information. The reinforcement learning method proposed in \cite{heess2017emergence} for example can achieve robust locomotion in rich environments by using very simple reward functions. Another promising method of generating locomotion behaviors aims at optimizing a sequence of end-effector trajectories through contact-invariant optimizations \cite{mordatch2012discovery}. This approach can produce realistic walking gaits \cite{posa2014direct}. However, a considerable off-line optimization effort is needed to obtain a single walking gait.

\subsection{The Proposed Method}
Although physics-based simulations can potentially produce various kinds of locomotion scenarios \cite{vaillant2017multi}, the control algorithm remains a big challenge. Even for very simple walking behaviors, these simulations can hardly go faster than real-time \cite{mordatch2010robust}. However, since dynamic equations are being integrated, interactions with the environment are made possible given stable controllers. Direct integration of multi-body symbolic equations \cite{docquier2013robotran} would slightly speed up the animation \cite{van2015biped}, but the effect of interaction forces should be included in the symbolic equations. A much faster speed can be achieved via interpolation or trajectory synthesis methods, however, producing interactions is not possible. In this work, we propose a method that can cover a wide range of walking conditions generated by physics-based simulations \cite{yin2007simbicon, coros2010generalized, mordatch2010robust} while offering 2-3 orders of magnitude faster simulation speeds. We use symbolic equations of a linear simplified model (3LP) \cite{faraji20173lp} that has closed-form solutions. While physics-based simulations need sub-millisecond integration times and hardly reach real-time factors, we use closed-form solutions of 3LP (as fast as microseconds only) to update the state only at display frames (e.g., $30$ frames per second). This boost of speed easily makes real-time crowd walking simulations possible. It also enables computationally limited or portable electronic devices to simulate interactive walking scenarios easily. 

To be precise, the proposed method is a novel hybrid combination of physics-based simulations and interpolation methods summarized in Fig.\ref{fig::architecture}. The physics of walking in our method is encoded in the 3LP model which is composed of three linear pendulums to model falling, swing and torso balancing dynamics. 3LP supports walking at different speeds, frequencies, double-support times, torso bending styles, terrain inclinations and subject heights and weights. Our symbolic equations also support external forces and torques applied to the torso while a previously developed controller \cite{faraji20173lp2} automatically captures these perturbations by adjusting footstep locations. This controller supports all the previously mentioned walking conditions without any parameter tuning. Since masses in the 3LP model are fixed to constant-height planes, we introduce a kinematic conversion to produce height variations. This part of our method involves adaptive trajectory synthesis without any tuning of parameters unlike the literature \cite{boulic1990global, li2000simulating}. Given a 3LP state (pelvis, torso and toe positions), our conversion adaptively varies the pelvis height to produce human-like excursions \cite{gard2004comparison}, lifts the swing toe to provide ground clearance \cite{wu2016determinants} and resolves a single Degree of Freedom (DoF) in each leg to produce thigh-shank-foot kinematics. These variations are all adaptive, independent of the previously mentioned walking conditions and calculated in closed-form.

\subsection{Novelties}
While covering a wide range of walking conditions, the proposed method simplifies physics-based simulations favoring faster computations. The 3LP model \cite{faraji20173lp} and the time-projection control \cite{faraji20173lp2} are originally developed to control a real robot for walking and push recovery applications \cite{faraji20173lp3}. Therefore, the novelty of this work mainly lies in adding torso styles and terrain inclination features to the 3LP model and more importantly, introducing an adaptive kinematic conversion to produce human-like gaits from 3LP states. This work mainly focuses on producing lower-limb kinematic trajectories while upper-body and pelvis oscillations can be included via predefined scalable trajectories similar to \cite{boulic1990global,li2000simulating}. An essential advantage of the 3LP model and time-projection control is in closed-form future predictions, given walking speed and external disturbance profiles. The proposed kinematic conversion method also produces human-like postures while preserving this property. In other words, the current converted posture does not depend on the previously converted postures. The entire method, therefore, enables a fast approximation of future kinematics in few microseconds which makes it suitable for model predictive control too. In the next two sections, we briefly introduce the 3LP model and the time-projection control scheme. Next, we formulate our adaptive kinematic conversion method. We continue the paper by demonstrating different walking gaits and conclude by a discussion on the supported range of walking conditions as well as promising aspects for future work.

\section{3LP Model}
\begin{figure}[]
    \centering
    \includegraphics[trim = 0mm 0mm 0mm 0mm, clip, width=0.45\textwidth]{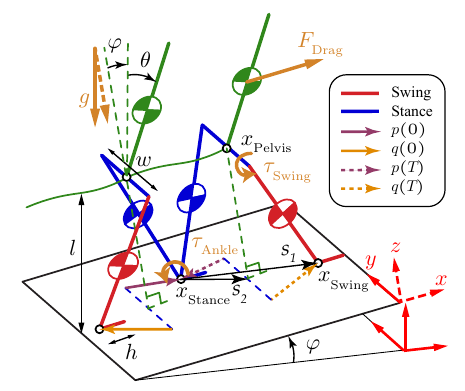}
    \caption{A schematic of the 3LP model used in this paper for walking gait generation \cite{faraji20173lp}. This 3D model is composed of three linear pendulums connected with a massless pelvis to simulate natural lateral bounces. All masses and the pelvis stay in constant-height planes which make the model linear. The torso also remains upright by an ideal actuator placed in the stance hip joint. The state of this model is described by the pelvis and swing foot horizontal positions while the swing hip and stance ankle torques serve as inputs to generate different walking gaits.} 
    \label{fig::3lp}
\end{figure}

The 3LP model \cite{faraji20173lp} is composed of three linear pendulums to simulate falling, swing and torso dynamics in walking. These pendulums are connected with a mass-less rigid pelvis which stays in a constant-height plane, similar to the LIP model. Each pendulum approximates a limb with a point mass in the middle and an inertia in the sagittal and lateral planes, but not around the pendulum rod itself. All masses stay in constant-height planes, assuming ideal prismatic actuators in the legs like the LIP model. Apart from the prismatic actuators, 3LP has three groups of two rotary actuators in lateral and sagittal directions, placed in the stance hip, swing hip and the stance ankle joints. This model simulates no pelvic and torso rotation, assuming ideal stance hip actuators to compensate the internal coupling between the limbs. The role of this actuator is to regulate the torso orientation similar to the control rules used in \cite{hodgins1997adapting, yin2007simbicon, coros2010generalized, mordatch2010robust}. Figure \ref{fig::3lp} demonstrates the 3LP model conceptually with the three pendulums, limb masses, state variables and actuation dimensions.

\subsection{Mechanics}
In this article, we skip complicated mechanical equations and refer to the original paper \cite{faraji20173lp} where the 3LP model was introduced. In brief, Newtonian equations of motion are written for each limb (and the pelvis) with incoming and outgoing forces and torques. Swing hip and stance ankle actuators are used as inputs. We consider the torques in these actuators as free variables while all other internal forces and torques are resolved by a symbolic combination of mechanical equations. In this paper, we only use swing hip torques for active control and leave the stance ankle torque profiles fixed, i.e., moving the CoP linearly from the heel to the toe or vice versa, depending on the direction of motion. The model state in 3LP is composed of horizontal pelvis and feet positions which together form a vector $x(t) \in \mathbb{R}^6$. All variables are expressed in a rotated coordinate frame attached to the slope, shown by dashes in Fig.\ref{fig::3lp}. The input vector $u(t) \in \mathbb{R}^4$ also represents swing hip and stance ankle torques. Finally, we define a constant vector $v \in \mathbb{R}^4$ defined as:
\begin{eqnarray}
x(t) = \begin{bmatrix} x_{Swing}(t) \\ x_{Pelvis}(t) \\ x_{Stance}(t) \end{bmatrix},\ \ u(t) = \begin{bmatrix} \tau_{Swing}(t) \\ \tau_{Ankle}(t) \end{bmatrix},\ \ v = \begin{bmatrix}
d \\ \sin(\theta+\phi) \\ \sin(\phi) \\ F_{Drag} \end{bmatrix}
\label{eqn::definitions}
\end{eqnarray}
which contains the variable $d=\pm1$ to determine left or right support phase, the fixed sines of torso bending and inclination angles ($\theta$ and $\phi$ respectively), and a constant sagittal dragging force $F_{Drag}$. As discussed later, these constant terms slightly change the periodic gaits, but not the derivation of control rules. All variables are depicted in Figure \ref{fig::3lp}. The overall equations of motion are found as:
\begin{eqnarray}
\frac{d^2}{dt^2}{x}(t) = C_x \ x(t) + C_u \ u(t) + C_v \ v
\label{eqn::continuous_3lp}
\end{eqnarray}
Inclusion of the variable $d$ is only because of a non-zero pelvis width to distinguishes between left and right bouncing. Note that the stance foot is fixed in 3LP and cannot slip. Therefore, the matrix $C_x$ does not influence $x_{Stance}(t)$ and its derivative (which is equal to zero), ensuring $\ddot{x}_{Stance}(t) = 0$ during the swing phase. After each step of motion, the swing and stance feet can be exchanged through multiplying $x(t)$ by the following matrix:
\begin{eqnarray}
S_x =& \begin{bmatrix}     \cdot & \cdot & I_{2\times2} \\
\cdot & I_{2\times2} & \cdot \\
I_{2\times2} & \cdot & \cdot 
\end{bmatrix}
\end{eqnarray}

In the original 3LP paper \cite{faraji20173lp}, we also derived equations for a double support phase by considering linear contact force transitions. In this case, the equation (\ref{eqn::continuous_3lp}) would keep both $\ddot{x}_{Stance}(t) = 0$ and $\ddot{x}_{Swing}(t) = 0$. The present paper adds torso bending and inclined walking features too. As mentioned earlier, these are fixed parameters in our model and we do not simulate changing slope or torso angles to avoid violating linearity assumptions.

\subsection{Closed-Form Equations}
By construction, 3LP equations are linear with respect to the state variables and inputs. To obtain closed-form equations for future predictions, it is enough to consider certain swing hip and ankle torque profiles and solve the second order differential equations of (\ref{eqn::continuous_3lp}) symbolically. We consider simple piecewise liner torque profiles which give us enough freedom for control and closely approximate human profiles \cite{faraji20173lp}:
\begin{eqnarray}
u(t) = u_{c} + t\ u_{r}
\end{eqnarray}
the parameters $u_{c}, u_{r} \in \mathbb{R}^4$ represent constant and time-increasing terms in both sagittal and lateral directions. With these definitions, closed-form 3LP equations are obtained as:
\begin{eqnarray}
q(t) = A(t) q(0) + B(t) u + C(t) v
\label{eqn::discrete_3lp}
\end{eqnarray}
where $q(t) \in \mathbb{R}^{12}$ and $u \in \mathbb{R}^8$ are:
\begin{eqnarray}
q(t) = \begin{bmatrix} x(t) \\ \dot{x}(t) \end{bmatrix} ,\ u = \begin{bmatrix} u_{c} \\ u_{r} \end{bmatrix}
\end{eqnarray}
and $0\le t \le T$. Considering a double support phase of a given duration $T_{ds}$, the single support phase will have a duration of $T_{ss} = T-T_{ds}$ where $T$ is the total step duration. Equations (\ref{eqn::discrete_3lp}) can be modified to consider the double support phase as well for any time $t \le T_{ds}$. We skip these derivations and refer to the original paper for details \cite{faraji20173lp}.

\subsection{Periodic Gaits}
Knowing the state evolution equations of 3LP in (\ref{eqn::discrete_3lp}), we can find successive phases of motion which are symmetric with respect to each other. Consider the matrix:
\begin{eqnarray}
M_x =& \begin{bmatrix}   I_{2\times2} & \cdot & -I_{2\times2} \\
\cdot & I_{2\times2} &-I_{2\times2} 
\end{bmatrix} 
\label{eqn::symmetry}
\end{eqnarray}
which extracts relative position vectors $s_{1}$ and $s_{2}$ from the vector $x(t)$, depicted in Figure \ref{fig::3lp}. At the beginning of each phase, these relative positions are the same in the sagittal direction and opposite in the lateral direction. Besides, the initial and final swing foot velocities together with the stance foot velocity should be zero (we assume impact-less locomotion in 3LP). Consider matrices $S$, $M$, $N$ and $O$ as:
\begin{eqnarray}
\nonumber S =& \begin{bmatrix} S_x & \cdot \\ \cdot & S_{x} \end{bmatrix}, \quad     M = \begin{bmatrix} M_x & \cdot \\ \cdot & M_{x} \end{bmatrix} \\
\nonumber N =& \begin{bmatrix}   \cdot & \cdot & \cdot & I_{2\times2} & \cdot & \cdot  \\ \cdot & \cdot & \cdot & \cdot & \cdot & I_{2\times2}  \end{bmatrix} \\
O =& diag(\ [1,-1, 1,-1, 1,-1, 1, -1]\ )
\end{eqnarray}
where $N$ implements initial foot velocity conditions and $O$ imposes gait symmetry. A valid periodic initial condition $\bar{Q}[k]$, inputs $\bar{U}[k]$ and constants $\bar{V}[k]$ must satisfy the following equations:
\begin{eqnarray}
\nonumber M \bar{Q}[k] &=& OMS(A(T)\bar{Q}[k]+B(T)\bar{U}[k]+C(T)\bar{V}[k]) \\
N \bar{Q}[k] &=& 0
\label{eqn::3lp_periodic}
\end{eqnarray}
which evolve $\bar{Q}[k]$ for one phase according to (\ref{eqn::discrete_3lp}), exchange the feet by $S$, extract the relative vectors $s_{1}$, $s_{2}$ and their derivatives by $M$ and apply the symmetry concept by $O$. The underlying assumption is a fixed gait frequency determined by the step time $T$. In this paper, we use capital letters to show discrete variables. The null-space formed by equations (\ref{eqn::3lp_periodic}) still leaves eight degrees of freedom. The lateral stance ankle torques are set to zero while the sagittal torques are set to a linear profile which together take four dimensions. The desired average forward velocity also takes one dimension while the remaining three dimensions are resolved by a minimization of hip torques. In other words, among all possible combinations of $\bar{Q}[k]$ and $\bar{U}[k]$ (for a fixed $\bar{V}[k]$) which form a null space of eight dimensions, we find a combination with the desired average speed, CoP profiles and minimal hip torques (refer to \cite{faraji20173lp} for further details). Therefore, open-loop 3LP gaits $\bar{q}(t)$ used in our controller are defined as:
\begin{eqnarray}
\bar{q}(t) = A(t) \bar{Q}[k] + B(t) \bar{U}[k] + C(t) \bar{V}[k]
\label{eqn::nominal_solution}
\end{eqnarray}
at each instance of time $0 \le t \le T$. Note that the double support phase is already encoded in matrices $A(t)$, $B(t)$ and $C(t)$ of equation (\ref{eqn::discrete_3lp}) as mentioned earlier.

\section{Time-Projection Control}

The 3LP model provides closed-form equations (\ref{eqn::discrete_3lp}) that describe system evolution over successive phases of motion. Thanks to linearity, it is very simple to find a Poincar\'e map or a linearized discrete model around the walking gait we obtained in the previous section. In this regard, we consider the same phase-change event, add an error to the state vector, evolve the erroneous state and a delta input until the end of the phase and measure the new error. Consider the stack of relative position vectors $s_1(t)$ and $s_2(t)$ and their derivatives which are extracted from the full vector by $s(t)=Mq(t)$ where $s(t) \in \mathbb{R}^8$. Consider also a matrix $\hat{M}$ defined as:
\begin{eqnarray}
\hat{M} =& \begin{bmatrix} \hat{M}_x & \cdot \\ \cdot & \hat{M}_x    \end{bmatrix}, \quad
\hat{M}_x =& \begin{bmatrix}   I_{2\times2} & \cdot \\
\cdot & I_{2\times2} \\ 
\cdot & \cdot \\ 
\end{bmatrix} 
\label{eqn::symmetry_reverse}
\end{eqnarray}
which adds an error $e(t)$ to the vector $s(t)$ and then to the full state vector $q(t)$. Since the relative vectors are extracted from the full vector by matrix $M$, the product $M \hat{M}$ naturally equals to identity. An initial error vector $E[k]$ added to $\bar{Q}[k]$ by $Q[k] = \hat{Q}[k]+\hat{M}E[k]$ evolves in time according to (\ref{eqn::discrete_3lp}) where an additional input $\Delta U[k]$ is also given to the system. The evolved error can be extracted from the next discrete state by:
\begin{eqnarray}
E[k+1] = OMS(Q[k+1] - \bar{Q}[k+1])
\label{eqn::error3}
\end{eqnarray} 
which leads to the following linear discrete model:
\begin{eqnarray}
\nonumber E[k+1] =& \hat{A}(T) E[k] + \hat{B}(T) \Delta U[k] \\
\hat{C} E[k+1] =& 0
\label{eqn::error_dynamics}
\end{eqnarray} 
where:
\begin{eqnarray}
\nonumber \hat{A}(T) =& O M S A(T) \hat{M} \\
\nonumber \hat{B}(T) =& O M S B(T) \\
\hat{C} =& \begin{bmatrix} 0_{2\times4} & I_{2\times2} & 0_{2\times2} \end{bmatrix}
\end{eqnarray}
and the matrix $\hat{C}$ is defined to constrain final swing foot velocities to zero.

\subsection{Discrete LQR Control}

Knowing error evolution equations (\ref{eqn::error_dynamics}) and the effect of inputs, we can design a Discrete Linear Quadratic Regulator (DLQR) controller to stabilize the system. Our particular system must satisfy a constraint on the foot velocity too which requires a simple manipulation of equations and DLQR's cost function. We refer to \cite{faraji20173lp2} for further details and only take the resulting control gain matrix $K$ which produces a corrective feedback $\Delta U[k] = -K E[k]$. A DLQR controller can be triggered at the beginning of each phase and produce a delta actuator input to correct the error. Due to a delayed reaction, however, this controller cannot reject intermittent disturbances optimally \cite{faraji20173lp2}. Therefore, we use a time-projecting controller that can react to inter-sample disturbances immediately by using the expertise of the DLQR controller. This idea is briefly introduced in the next section.

\subsection{Continuous Time-Projection}

\begin{figure*}[]
    \centering
    \includegraphics[trim = 0mm 0mm 0mm 0mm, clip, width=1\textwidth]{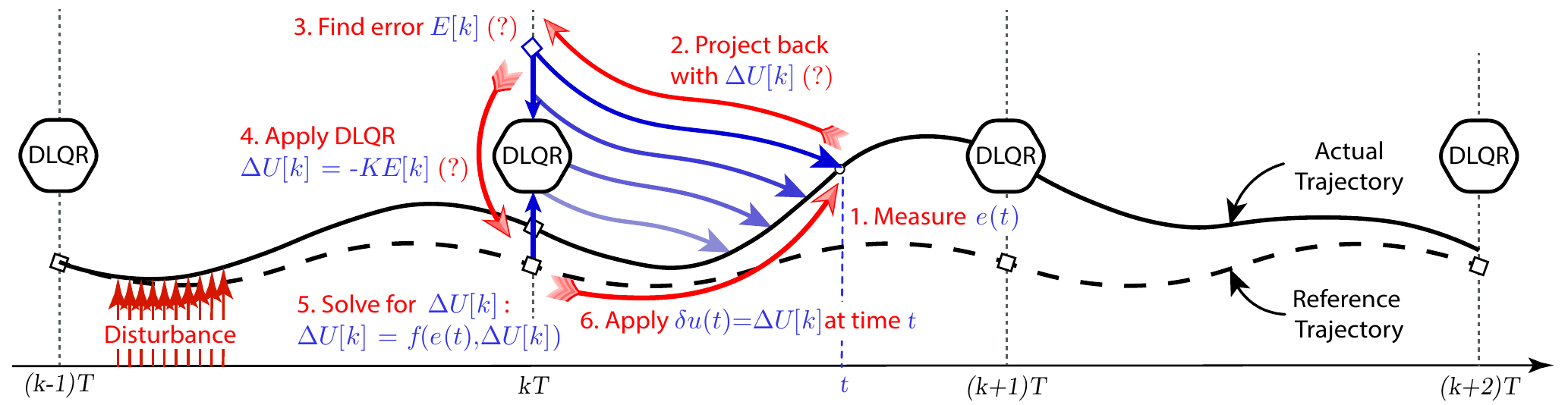}
    \caption{Time-projection control over a nominal system trajectory. Imagine a control period $T$ is decided for which, a DLQR controller is calculated to stabilize the system. For the walking application, we set $T$ equal to the actual step time. The DLQR controller can only be triggered at time instances $kT$ which makes it sensitive to inter-sample disturbances, especially if the system has largely unstable modes. However, we can take advantage of DLQR knowledge and provide immediate corrections at every inter-sample time $t$. Knowing system evolution matrices from the closest discrete sample $kT$ until the time $t$, we can project the measured error $e(t)$ back in time and obtain an equivalent discrete error $E[k]$. This discrete error can evolve in time with some $\Delta U[k]$ and lead to the currently observed continuous error. To resolve the ambiguity between $E[k]$ and $\Delta U[k]$, we link them together by the DLQR controller in step 4 of time-projection. Therefore, the DLQR controller provides a stabilizing input which we directly apply to the system at time $t$.} 
    \label{fig::projecting}
\end{figure*}

To formulate the time-projection idea, we consider a free system without constraints to present simpler formulations and provide an easier understanding. Handling time-projection for constrained systems can be found in the original control paper \cite{faraji20173lp2}. After solving 3LP equations in closed-form and deriving discrete error dynamics, we showed that we could easily find a DLQR controller that stabilizes the system by swing hip torque adjustments and in consequence, footstep adjustments. Now consider an on-line control paradigm in which we can measure the system error at any inter-sample time and react to it quickly. The reaction might not stabilize the system immediately like CoP modulation, but a proper adjustment of swing foot location can considerably save control effort in the following phases. This is because the current swing position (which is weakly coupled to other system variables) later becomes the new stance foot position which tightly couples to other system variables (mainly the CoM). Consider an inter-sample time $t$ where $0\le t \le T$, the nominal periodic solution $\bar{q}(t)$ defined in (\ref{eqn::nominal_solution}), the current measured state $q(t)$ and the instantaneous error:
\begin{eqnarray}
e(t) = M(q(t)-\bar{q}(t))
\label{eqn::current_error}
\end{eqnarray}
We can apply a corrective swing hip torque $\delta u(t)$ at the time $t$ in addition to the nominal actuator inputs $\bar{u}(t) = \bar{u}_c + t \bar{u}_r$ (we assume fixed CoP profiles in the stance foot and only use the swing hip torques for online control). This scenario is depicted in Fig.\ref{fig::projecting} in details. The time-projection controller takes the following steps in sequence to find the vector $\delta u(t)$:
\begin{enumerate}
    \item Measure the current numeric error $e(t)$ at time $t$. 
    \item Consider an unknown parametric input $\Delta U[k]$ until time $t$ and project the measured error $e(t)$ back in time.
    \item Calculate a possible parametric initial vector $E[k]$ by:
    \begin{eqnarray}
    e(t) = A(t-kT) E[k] + B(t-kT) \Delta U[k]
    \end{eqnarray}
    \item Apply DLQR on $E[k]$ to find $\Delta U[k]$:
    \begin{eqnarray}
    \Delta U[k] = -K E[k]
    \end{eqnarray}
    \item Now find $\Delta U[k]$ by solving a linear system of equations:
    \begin{eqnarray}
    \begin{bmatrix} A(t-kT) & B(t-kT) \\ K & I \end{bmatrix} \begin{bmatrix}
    E[k] \\ \Delta U[k] \end{bmatrix} = \begin{bmatrix} e(t) \\ \cdot
    \end{bmatrix}
    \label{eqn::simple_system_solvedU}
    \end{eqnarray}
    \item Assuming $\delta u(t) = \Delta U[k]$, take the resulting corrective input and apply it to the system at time $t$.
\end{enumerate}
The time-projection control involves solving a linear system of equations at every control sample which only takes few microseconds on a modern computer, compared to at least few hundred microseconds taken by our previous MPC controller introduced in \cite{faraji2014robust}. Note that the projecting controller does not need to know disturbance forces, and produces hip torques which result in footstep adjustments. In the absence of errors $e(t)$ also, this controller produces no correction. In the absence of disturbance forces, however,  while the error $e(t)$ is non-zero, it produces a constant $\delta u(t)$ which is equal to a discrete correction that the DLQR controller would produce at the beginning of the phase. In other words, as long as the system evolves without inter-sample disturbances, the projecting controller has no advantage over the DLQR controller in providing on-line corrections. 

In our previous work \cite{faraji20173lp2}, we extensively discuss the recovery performance of time-projection control against different inter-sample push timings. We also analyze controllable regions, given typical hardware limitations existing in bipedal systems. Due to an online numerical optimization, MPC can handle such inequality constraints easily. The time-projecting controller is blind to these constraints. However, our analysis in \cite{faraji20173lp2} shows that in normal walking conditions, the time-projection controller covers most of the whole set of controllable states. Therefore, MPC might only cover a slightly bigger set of states which are not visited most of the time in practice \cite{faraji20173lp2}. Having an approximate of stable human walking trajectories by the 3LP model and time-projection control, in the next section, we introduce a conversion to slightly modify these trajectories in the vertical direction to produce more human-like motions.

\section{Kinematic Conversion}
The idea of kinematic conversion is to make the 3LP posture more human-like. As mentioned earlier, we focus on adding pelvis vertical excursions, ground clearance and lower-limb motions to the gait. Other human-like features like pelvis rotations and upper body motions are not considered in the present work. Our simple conversion algorithm only needs the current 3LP state and phase time $t$ as inputs. Knowing other constants including phase timing parameters $T_{ds}$, $T_{ss}$, the slope angle $\phi$, anatomical properties and the ground clearance height, our conversion algorithm performs the following steps:
\begin{enumerate}
    \item Pelvis height: Find a smooth pelvis height trajectory based on relative feet and pelvis positions.
    \item Ground clearance: Calculate a simple vertical toe trajectory as a function of phase time $t$.
    \item Knee target points: Find two target points for the knees that translate to hip angles and solve the inverse kinematics redundancy of each leg.
\end{enumerate}
These small modifications are based on trajectory synthesis and minimally influence 3LP's overall falling and swing dynamics. Here, we simply rely on decoupling assumption between vertical and horizontal dynamics similar to \cite{mordatch2010robust, you2016straight}. A more precise but computationally expensive approach would be to use the full dynamics equations and perform vertical adjustments only within the null-space of horizontal tasks \cite{griffin2017straight}. The remainder of this section describes our adaptive kinematic conversion method in details. 

\subsection{Pelvis Height}
The horizontal dynamics of walking in our simulations is approximated by the 3LP model assuming constant heights for the pelvis, the two feet, and all limb masses. Given a 3LP state in periodic or transient walking conditions, our kinematic conversion takes the relative foot-hip positions and calculates a smooth pelvis height trajectory. This is done via a simple mixture of geometric variables at each instance of time. Remember that the legs in 3LP are modeled by extensible prismatic actuators. Given certain footstep locations, if we assume fixed-length legs like the normal inverted pendulum model, we obtain arc shapes for the pelvis which sharply intersect together \cite{kuo2005energetic}. Human pelvis trajectories are similar to smooth sine shapes, however, going to a minimum during the double support phase \cite{gard2004comparison}. To produce such trajectories, we use two different methods:  
\begin{enumerate}
    \item Fixed mixture: we consider fixed leg-length arcs around the CoP in each foot and introduce a soft weighting between them to produce the final pelvis height profile.
    \item Adaptive mixture: we consider variable leg-length arcs around the CoP in each foot and apply a soft minimum function to produce the final pelvis height profile.
\end{enumerate}
While the first method is enough for periodic walking and small perturbations, it cannot support backward walking or extreme toe-off stretching in perturbed conditions. The second method, however, simply produces a feasible pelvis height for both legs. 

\begin{figure*}[]
    \centering
    \includegraphics[trim = 0mm 0mm 0mm 0mm, clip, width=0.8\textwidth]{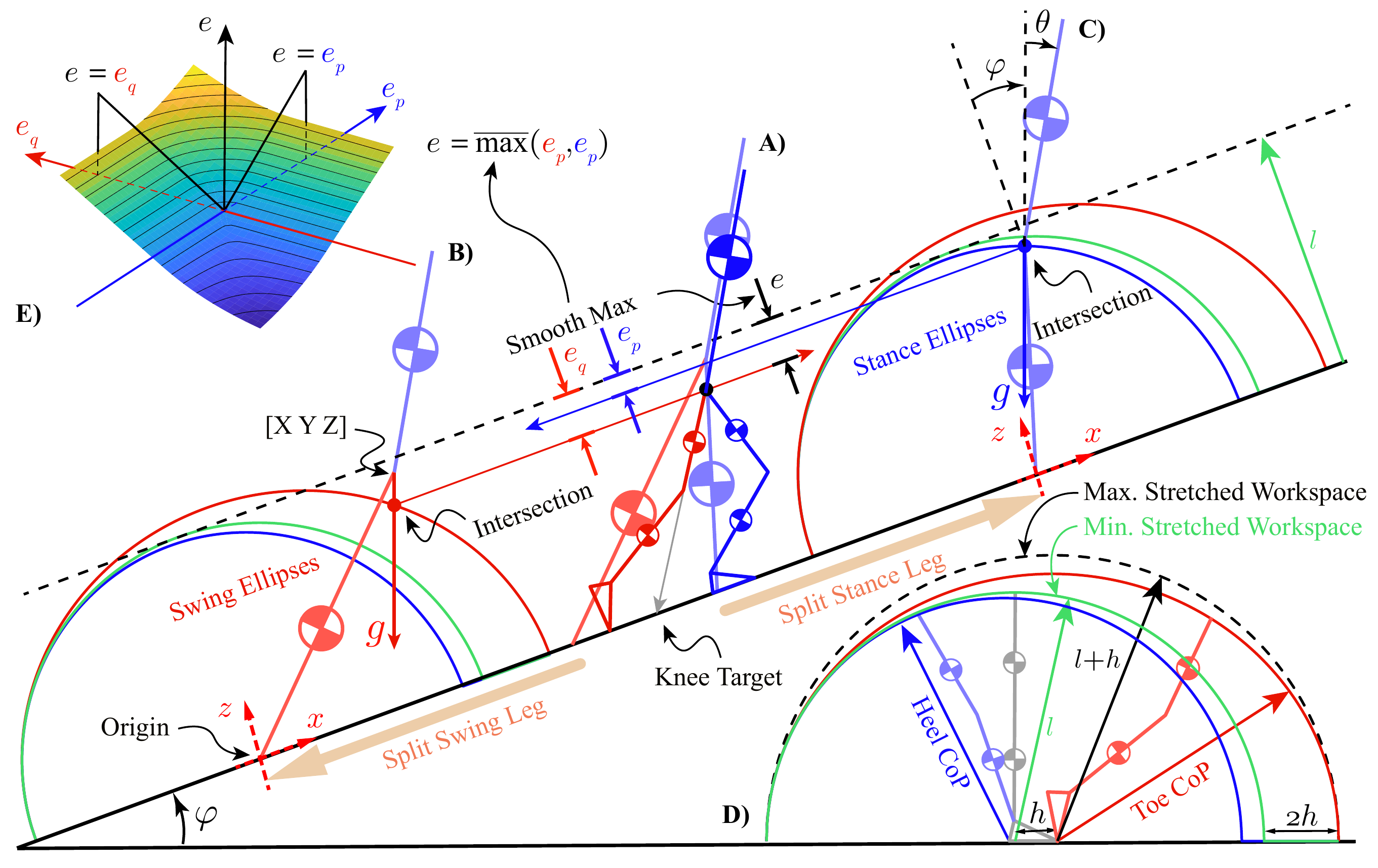}
    \caption{A snapshot of inclined walking in early swing phase. A) The given 3LP state in the background can be split into B) swing and C) stance legs. Depending on the sagittal distance of the two feet, we calculate CoP points (virtual in the swing foot) and create candidate pelvis height arcs around the heel points. For the trailing leg, the CoP is on the toe which produces the red arc. For the front leg, the CoP is on the heel which creates the blue arc. For each leg, by intersecting a vertical line (passing through the pelvis in 3LP) with the arc, we find a candidate pelvis height. Now, compared to the maximum height possible (equals the leg length $l$), we calculate errors $e_p$ and $e_q$ for each leg and apply a smooth maximum function on them which is shown in E). The resulting maximum $e$ determines the final pelvis height with respect to the maximum height $l$. Given the pelvis and toe positions for each leg then, we just need to resolve a single degree of freedom in each leg to find the complete thigh-shank-foot posture. This is done by determining a target point for the knee on the ground which implicitly determines the desired hip angle. In case the foot penetrates the ground, this angle is adjusted to keep the foot always flat.} 
    \label{fig::onephase}
\end{figure*}

Remember that all the 3LP equations are solved in a rotated coordinate frame attached to the slope (refer to Fig.\ref{fig::onephase}). Therefore, the gravity vector is rotated and the nominal leg length (the variable $l$ in Fig.\ref{fig::3lp}) is reduced by a factor of $\cos(\phi)$. This ensures a stretched leg when walking at zero speed on the slope. For a given 3LP state, we define relative vectors $p(t)$ and $q(t)$ by:
\begin{eqnarray}
\nonumber p(t) &=& P\ (x_{Stance}(t)-x_{Pelvis}(t))+ \begin{bmatrix}
0 & \frac{wd}{2}\end{bmatrix}^T\\ q(t) &=& P\ (x_{Swing}(t)-x_{Pelvis}(t))+ \begin{bmatrix}
0 & -\frac{wd}{2}\end{bmatrix}^T
\label{eqn::pq}
\end{eqnarray}
where the variable $d=\pm1$ indicates left or right support phases (used in (\ref{eqn::definitions})), the parameter $w$ denotes pelvis width and the matrix $P$ is a simple operator to project the quantities on the $x-y$ plane shown in Fig.\ref{fig::3lp}. Example sagittal components of $p(t)$ and $q(t)$ are shown in Fig.\ref{fig::pelvisheight}E for an adult person (height of $1.7m$) walking at a speed of $1m/s$ and a frequency of $1.7\ steps/s$. During in-place walking, we assume that the CoPs stay in the middle of each foot (the CoP in swing foot is virtual). Depending on the distance between the two feet, we move CoPs to the toes or heels. This is done via a linear profile in which a step length equal to the leg length produces a half-foot CoP movement:
\begin{eqnarray}
\nonumber p_{CoP}(t) &=& \frac{h}{2} (1+\frac{q(t)-p(t)}{l})\\ 
q_{CoP}(t) &=& \frac{h}{2} (1+\frac{p(t)-q(t)}{l})
\label{eqn::pqcop}
\end{eqnarray}
where the parameters $h$ and $l$ denote foot length and leg length shown in Fig.\ref{fig::3lp} respectively. We also assume no lateral movements for the CoP in each foot. As observed in Fig.\ref{fig::pelvisheight}E, the trajectories $p(t)$ and $q(t)$ have non-zero derivatives at the boundary times (in the beginning and at the end of the step phase). These derivatives produce sharply intersecting arc shapes for the pelvis whereas setting them to zero would produce flat trajectories. To achieve this, we define an additional signal $\alpha(t)\dot{p}(t)$ where the polynomial $\alpha(t)$ has derivatives of $-1$ at $t=T_{ds}$ and $t=T$ shown in Fig.\ref{fig::pelvisheight}A. This signal can correct the derivatives by setting them to zero when added to $p(t)$ and $q(t)$ (shown in Fig.\ref{fig::pelvisheight}E). Therefore, modified relative positions $\bar{p}(t)$ and $\bar{q}(t)$ can be defined as:
\begin{eqnarray}
\nonumber \bar{p}(t) &=& p(t) + \alpha(t)\dot{p}(t)\\ 
          \bar{q}(t) &=& q(t) + \alpha(t)\dot{p}(t)
\label{eqn::pqbar}
\end{eqnarray}
which represent smooth relative positions between the hip and the heel in each leg, ensuring zero derivatives at the phase boundary times. Assume we put a rotated coordinate frame (along the slope) on each heel and express the modified relative 3LP pelvis position by $[X\ Y\ Z]$ in this frame where $Z=l\cos(\phi)$ shown in Fig.\ref{fig::onephase}B. Now, fixed leg-length arcs can be found as:
\begin{eqnarray}
Z^2 = (X + x_{CoP} +l\sin(\phi))^2 + Y^2 + z_{fixed}(X,Y,x_{CoP})^2
\label{eqn::zfixed}
\end{eqnarray}
where the constant $l\sin(\phi)$ is added to compensate the effect of slope. The candidate pelvis heights for each leg are:
\begin{eqnarray}
\nonumber z_{p,fixed}(t) =& z_{fixed}(\bar{p}_x(t),\bar{p}_y(t),p_{CoP}(t)) \\
z_{q,fixed}(t) =& z_{fixed}(\bar{q}_x(t),\bar{q}_y(t),q_{CoP}(t))
\label{eqn::zpqfixed}
\end{eqnarray}
where $x$ and $y$ are sagittal and lateral components respectively. Now, according to the first method, the final pelvis height trajectory could be found by a smooth transition from the arc on the stance foot $z_{p,fixed}(t)$ to the arc on the swing foot $z_{q,fixed}(t)$:
\begin{eqnarray}
z_{fixed}(t) = (1-\gamma(t))\ z_{p,fixed}(t) + \gamma(t)\ z_{q,fixed}(t)
\label{eqn::zpelvis}
\end{eqnarray}
which is shown in Fig.\ref{fig::pelvisheight}F. The function $\gamma(t)$ implements a smooth transition shown in Fig.\ref{fig::pelvisheight}C. The resulting curve $z_{fixed}(t)$ features zero derivatives at the phase boundary times while it peaks approximately in the middle of the single support phase. Overall, the converted pelvis position is matching 3LP's pelvis position horizontally and shifted down from $l\cos(\phi)$ to the new height $z_{fixed}(t)$. The first method produces a good approximate of human trajectories, but there is no guaranty that the next touch-down happens on the heel. In backward walking or perturbed conditions, for example, the next step might touch down on the toes. In these cases, the resulting pelvis height (at $t=T$) should be feasible for the other stance leg which is still on the heel. However, the formula (\ref{eqn::zpelvis}) always converges to the swing arc at $t=T$ which could become infeasible. Therefore, the mixture used in (\ref{eqn::zpelvis}) is meaningful in terms of producing vertical excursions, but limited to periodic forward walking conditions.

To overcome limitations of the first method, we replace the time-based mixture of (\ref{eqn::zpelvis}) with a smooth maximum function in the second method to better decide between swing and stance leg arcs. Also, we slightly modify the arcs to allow for more leg flexion and extension in the stance phase. Consider Fig.\ref{fig::onephase}D which shows different pelvis arcs depending on the position of CoP. When walking in-place, no matter what other gait parameters are, the arc is calculated around the heel position. In maximum step length conditions, however, the CoP moves to extremities according to (\ref{eqn::pqcop}). In this case, we use a specific nonlinear function $\delta(x)$ (Fig.\ref{fig::pelvisheight}D) which maps the linear CoP movements of (\ref{eqn::pqcop}) into asymmetric profiles:
\begin{eqnarray}
f(x_{CoP}) = \Delta = \frac{h}{2} \delta(\frac{2}{h}\ x_{CoP}-1)
\label{eqn::cop_asym}
\end{eqnarray}
When the CoP goes to the toes ($x_{CoP}=h$), the function $f$ produces a value of $h/2$ and when the CoP goes to the heels ($x_{CoP}=0$), this function returns $-\epsilon h/2$. Based on this function, we formulate our modified arcs by following ellipses:
\begin{eqnarray}
(\frac{x-2\Delta}{l+2\Delta})^2 + (\frac{y}{l})^2 + (\frac{z}{l+\Delta}) = 1
\label{eqn::ellipses}
\end{eqnarray}
which are shown in Fig.\ref{fig::onephase}D. The dashed circle in this plot shows the maximum workspace of the pelvis rotating around the toe. We consider smaller arcs (the ellipse shown in red) when the CoP is at the toes to avoid extra lifting and over-extension. The green circle also shows the minimum workspace when rotating around the heel (for in-place walking). When the CoP is at the heel, the blue ellipse produces a small knee flexion at the touch-down moment (like human \cite{delp2008muscle}) determined by the choice of $\epsilon=0.2$ shown in Fig.\ref{fig::onephase}D and Fig.\ref{fig::pelvisheight}D. Our specific design of $\delta(x)$ and adaptive ellipses of (\ref{eqn::ellipses}) produce convincing human-like trajectories compared to a few human gaits recorded (discussed in the next section). However, they could be tuned further in future work to better match human trajectories in a wider range of walking conditions. 

Given a 3LP state, we calculate the CoP points, split the legs and create an ellipse for each of them in the rotated coordinate frame (shown in Fig.\ref{fig::onephase}). A vertical line coming down from the pelvis in each case intersects with the ellipse and determines the candidate pelvis height of that leg. Denoting the modified relative 3LP pelvis position by $[X\ Y\ Z]$ like before, the vertical line in the rotated coordinate frame is:
\begin{eqnarray}
y=Y , \ x-X = \tan(\phi) (z-Z)
\label{eqn::vertical_line}
\end{eqnarray}
which intersects with the ellipse of (\ref{eqn::ellipses}) and results in the following equation (as a function of $z$):
\begin{eqnarray}
(\frac{X+\tan(\phi)(z-Z)-2\Delta}{l+2\Delta})^2 + (\frac{Y}{l})^2 + (\frac{z}{l+\Delta}) = 1
\label{eqn::intersection}
\end{eqnarray}
The height of intersection point (denoted by $z=z_{adapt}(X,Y,x_{CoP})$) solves the equation (\ref{eqn::intersection}) and therefore, each leg gives a candidate pelvis height:
\begin{eqnarray}
\nonumber z_{p,adapt}(t) =& z_{adapt}(\bar{p}_x(t),\bar{p}_y(t),p_{CoP}(t)) \\
z_{q,adapt}(t) =& z_{adapt}(\bar{q}_x(t),\bar{q}_y(t),q_{CoP}(t))
\label{eqn::zadaptive}
\end{eqnarray}
Unlike the time-based mixture of (\ref{eqn::zpelvis}), in the second method, we use a smooth minimum function between $z_{p,adapt}(t)$ and $z_{q,adapt}(t)$ which is implemented as:
\begin{eqnarray}
z_{adapt}(t) = l - \overline{\max}(l-z_{p,adapt}(t),\ l-z_{z,adapt}(t))
\label{eqn::smoothmin}
\end{eqnarray}
where $\overline{\max}(a,b)$ is defined as:
\begin{eqnarray}
\overline{\max}(a,b) = \left\{
\begin{array}{ll}
\sqrt{a^2+b^2} & 0 \le a,b\\
a & b<0\le a \\
b & a<0\le b \\
a + b + \sqrt{a^2+b^2} & a,b <0 \\
\end{array}
\right.
\label{eqn::smoothmax}
\end{eqnarray}
and shown in Fig.\ref{fig::onephase}E. The time-trajectory of $z_{adapt}(t)$ shown in Fig.\ref{fig::pelvisheight}G is very similar to $z_{fixed}(t)$ (shown in Fig.\ref{fig::pelvisheight}F) in periodic walking conditions. Fig.\ref{fig::onephase}A also visualizes the mechanism of finding individual pelvis heights and the smooth maximum function $\overline{\max}(a,b)$. Due to the fact that we always choose the minimum pelvis height in the second method (which is of course feasible for both legs), we can support non-periodic and backward walking as well as extremely asymmetric triangular leg coordinations which happen in inclined walking or presence of dragging forces. Therefore, we prefer the second adaptive method over the first fixed-time mixture method. The modified ellipsoid design of the arc shapes also produces realistic knee flexion during the touch-down and ankle extension during the push-off moments.

\begin{figure}[]
	\centering
	\includegraphics[trim = 0mm 0mm 0mm 0mm, clip, width=0.47\textwidth]{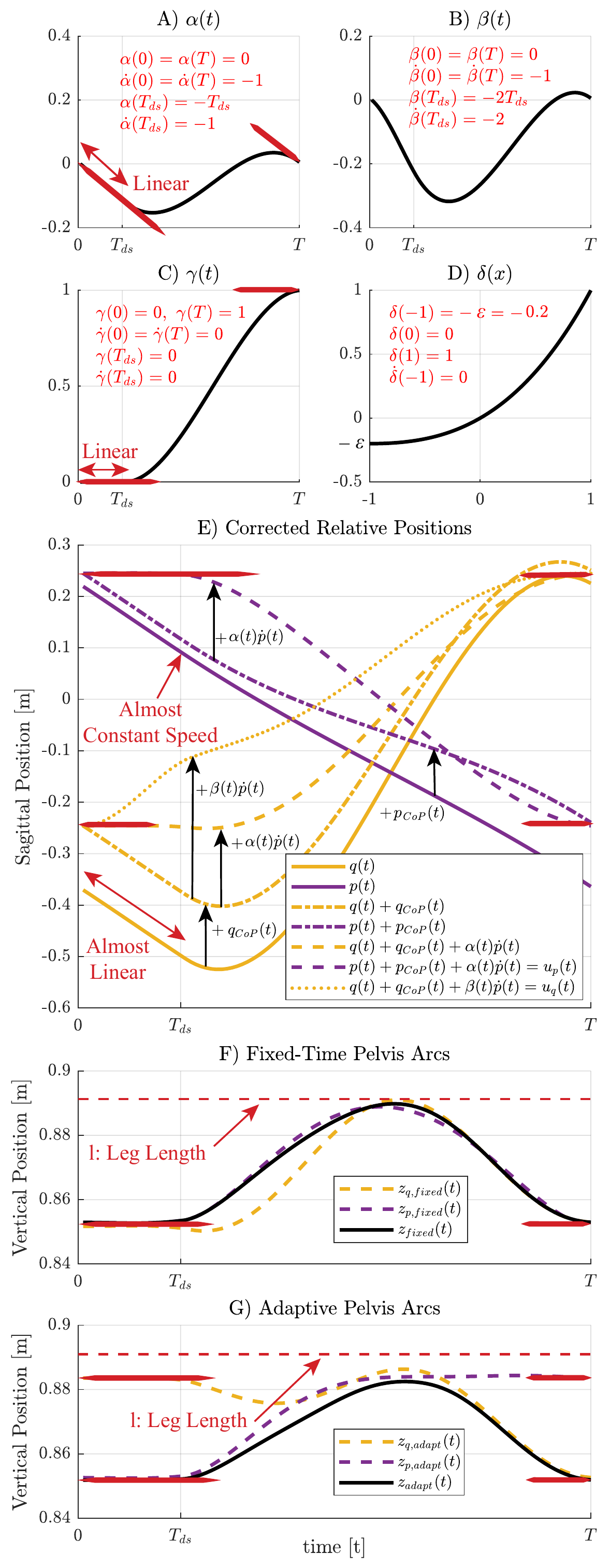}
\end{figure}
\begin{figure} []
	\caption{Composition of smooth relative positions and knee target trajectories based on relative foot positions and phase timing of 3LP. A) The polynomial $\alpha(t)$ is used to correct the nonzero velocities of $p(t)$ and $q(t)$ at the boundary times. When multiplied by $\dot{p}(t)$ in and added to $p(t)$ or $q(t)$, the specific design of $\alpha(t)$ can result in zero derivatives at the boundaries. B) The polynomial trajectory $\beta(t)$ is designed to perform a faster progress than $\alpha(t)$. When polynomial $\beta(t)$ is multiplied by $\dot{p}(t)$ in and added to $q(t)$, it produces a trajectory which moves forward even during $0\le t \le T_{ds}$. C) The polynomial trajectory $\gamma(t)$ used for a time-based mixture of $z_{p,fixed}(t)$ and $z_{q,fixed}(t)$ in the first method to produce a soft transition. D) The nonlinear function $\delta(x)$ which maps the linear motions of CoPs into asymmetric profiles $\Delta$ for each leg. These profiles adjust the default circular pelvis arcs (shown in green, Fig.\ref{fig::onephase}D) into different ellipses for swing and stance legs according to (\ref{eqn::ellipses}). E) Original relative foot-hip positions $p(t)$ and $q(t)$, addition of smooth CoP trajectories $p_{CoP}(t)$ and $q_{CoP}(t)$ and addition of corrective signals $\alpha(t)\dot{p}(t)$ and $\beta(t)\dot{p}(t)$ to produce smooth trajectories $\bar{q}(t)$ and $\bar{p}(t)$ (used for pelvis height trajectory generation) and $u_q(t)$ (used together with $u_p(t)$ as knee target trajectories). The synthesized signals are mere functions of 3LP state and phase timing without any history or dependency on the past. F) The arc trajectories $z_{p,fixed}(t)$ and $z_{q,fixed}(t)$ (produced from $\bar{p}(t)$,  $\bar{q}(t)$, $p_{CoP}(t)$ and $q_{CoP}(t)$) are smoothly combined together with $\gamma(t)$ to generate the final pelvis height trajectory $z_{fixed}(t)$. G) The adaptive arc trajectories $z_{p,adapt}(t)$ and $z_{q,adapt}(t)$ are smoothly combined together by $\overline{\max}(a,b)$ to generate the final adaptive pelvis height trajectory $z_{adapt}(t)$.}
	\label{fig::pelvisheight}
\end{figure}

\subsection{Ground Clearance}
Remember that both feet in the 3LP model are constrained to have a zero height. To make them more realistic, we consider simple vertical sinusoid curves that lift the swing toe vertically. These curves are scaled by the ground clearance parameter as a percentage of the leg length. Our simple design of these trajectories produces realistic motions, but cannot simulate foot flapping effects shortly after the touch-down moment (in which the foot completely lands on the ground after the heel-strike). We consider adding this feature in future work.

\subsection{Knee Target Points}
Given the pelvis and toe positions in the Cartesian space, the task in this stage is to resolve a single degree of freedom in each leg to find a human-like thigh-shank-foot posture. Our strategy is to determine the hip angle based on certain target trajectories on the ground. In each leg, the thigh vector (connecting the hip to the knee joint) points towards a target trajectory on the ground shown in Fig.\ref{fig::onephase}A. Once the hip angle is determined, the configurations of shank and foot segments are found by solving a simple Inverse Kinematic (IK) problem between the knee and the toe, restricting the foot segment inside the sagittal plane. Our IK formulation also does not allow for a heel position below the toe position vertically. 

\begin{figure}[]
    \centering
    \includegraphics[trim = 0mm 0mm 0mm 0mm, clip, width=0.48\textwidth]{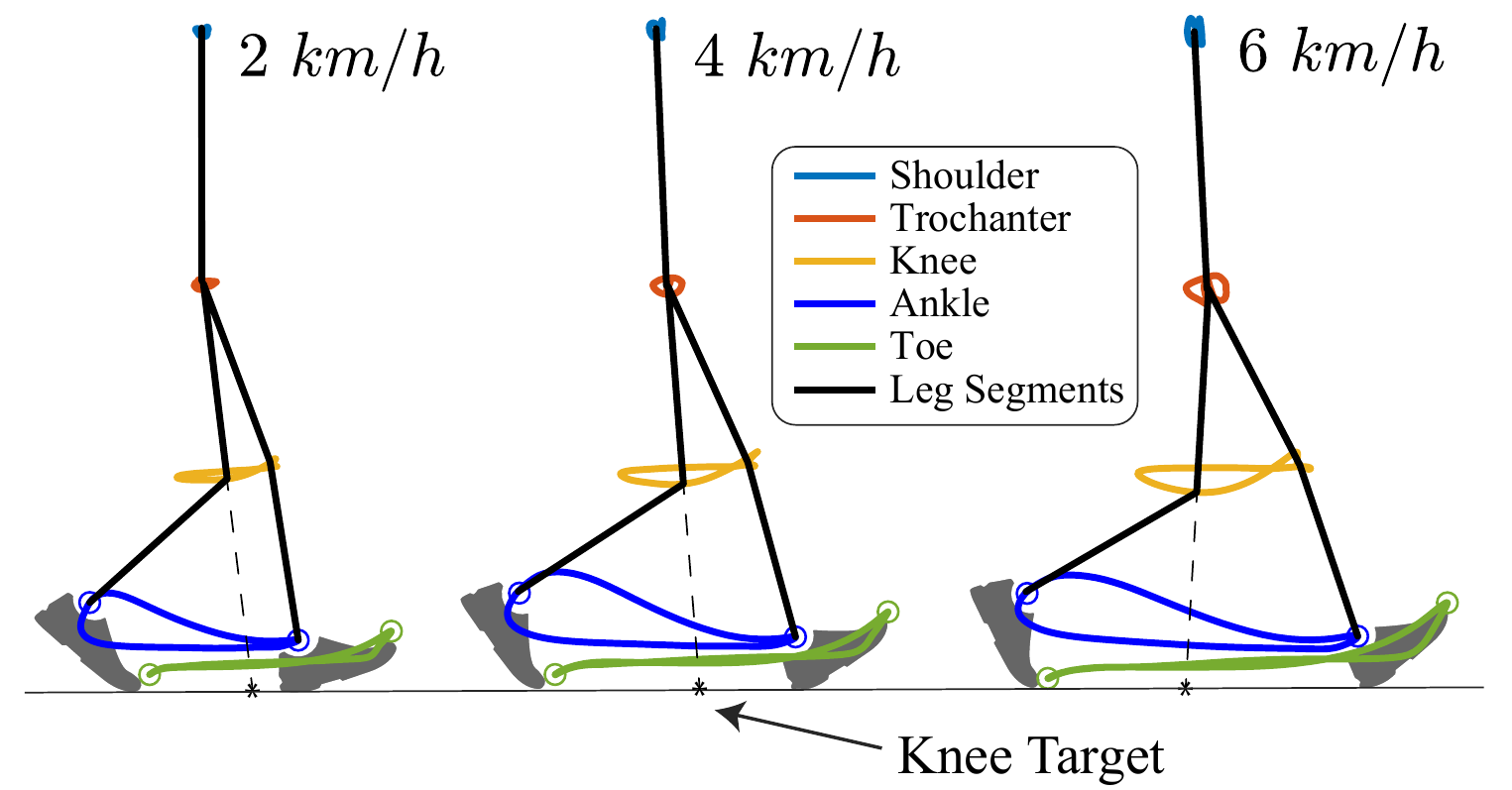}
    \caption{Kinematic configurations of the leg segments in human walking at different speeds (subject height of $1.76m$). The black segments indicate body postures at the touch-down and toe-off moments approximately. Given the knee marker trajectories, it is obvious that the thigh segment already starts swinging forward before the entire leg starts its swing phase. This requires target swing trajectories on the ground that progress forward during the double support phase already.} 
    \label{fig::kneetraj}
\end{figure}

We design the knee target trajectories by a similar mixture of geometric variables (foot positions) introduced earlier. In fact, for the stance leg, the relative position:
\begin{eqnarray}
u_p(t) = p(t) + \alpha(t)\dot{p}(t) + p_{CoP}(t)
\label{eqn::u_p}
\end{eqnarray}
is a good target trajectory. A careful inspection of human trajectories at different walking speeds reveals that the Cartesian swing knee position already starts moving forward before the swing phase starts. This effect is shown in Fig.\ref{fig::kneetraj} at different walking speeds. The relative position $q(t)+ \alpha(t)\dot{p}(t)+q_{CoP}(t)$ is not a good target trajectory for the swing leg, since it remains constant during the double support phase and starts moving forward only in the single support phase (Fig.\ref{fig::pelvisheight}E). To overcome this issue, we introduce a polynomial $\beta(t)$ shown in Fig.\ref{fig::pelvisheight}B, a modified version of $\alpha(t)$ which is approximately two times larger, but with similar derivative properties. The swing target point is now defined by:
\begin{eqnarray}
u_q(t) = q(t) + \beta(t)\dot{p}(t) + q_{CoP}(t)
\label{eqn::u_q}
\end{eqnarray}
which already starts moving forward during the double support phase shown in Fig.\ref{fig::pelvisheight}E. The two target trajectories $u_p(t)$ and $u_q(t)$ are eventually used to find the hip angles which then determine the leg configuration completely.

\section{Results}

\begin{figure*}[]
    \centering
    \includegraphics[trim = 0mm 0mm 0mm 0mm, clip, width=0.8\textwidth]{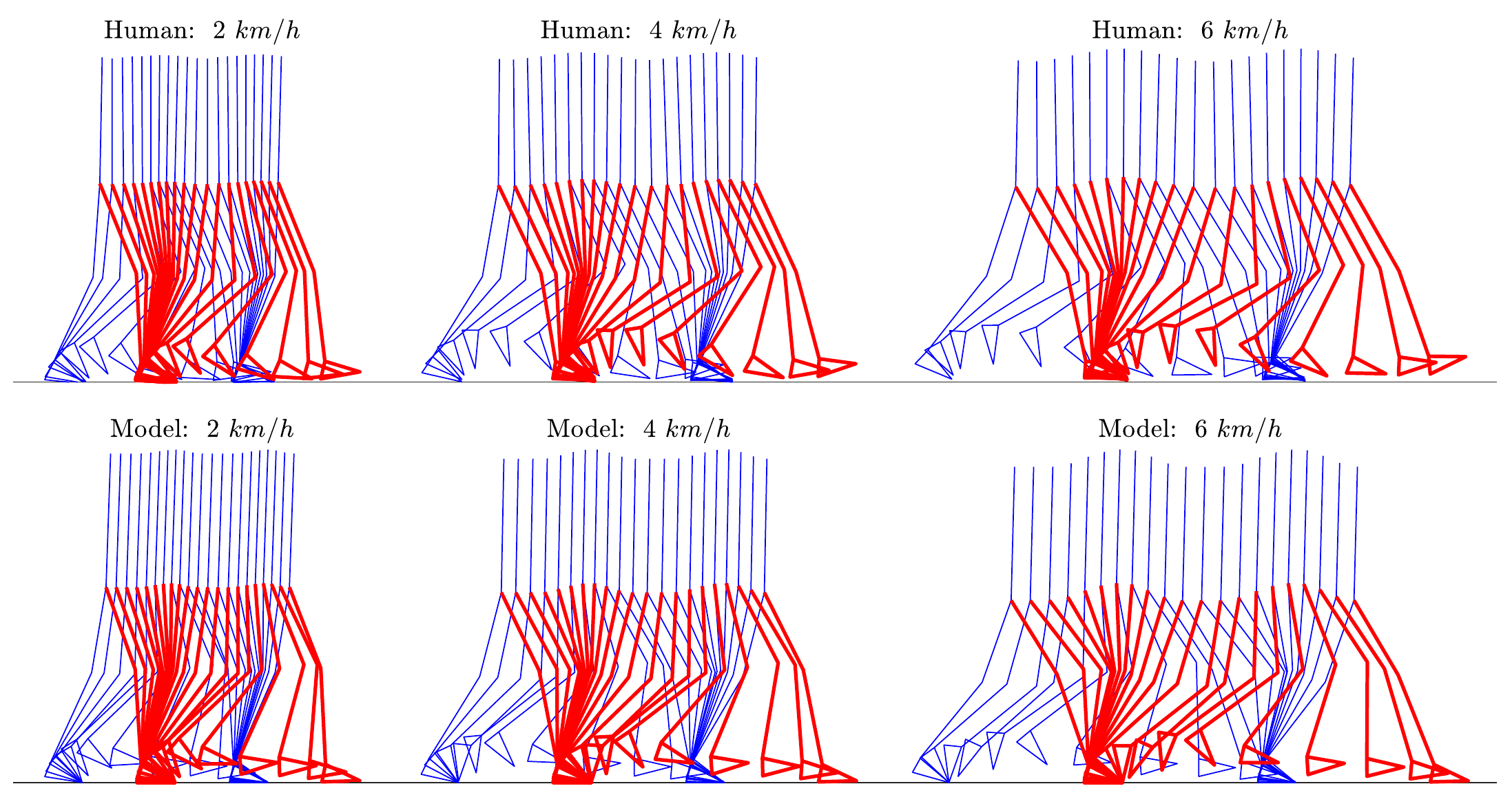}
    \caption{Snapshots of human walking and synthesized walking trajectories at different speeds. The choice of step frequency, double support duration, torso angle and ground clearance parameters as well as body properties are taken from each human experiment and used in the corresponding simulation. The overall horizontal dynamics of walking is encoded in the 3LP model which produces human-like limb motions. On top of 3LP trajectories, our kinematic conversion can produce vertical excursions for the pelvis, human-like knee angles at the touch-down moments and realistic coordinations of thigh-shank-foot segments. However, the current method is unable to produce foot flapping motions after the heel-strike.}
    \label{fig::plot_speeds}
\end{figure*}

\begin{figure*}[]
	\centering
	\includegraphics[trim = 0mm 0mm 0mm 0mm, clip, width=0.8\textwidth]{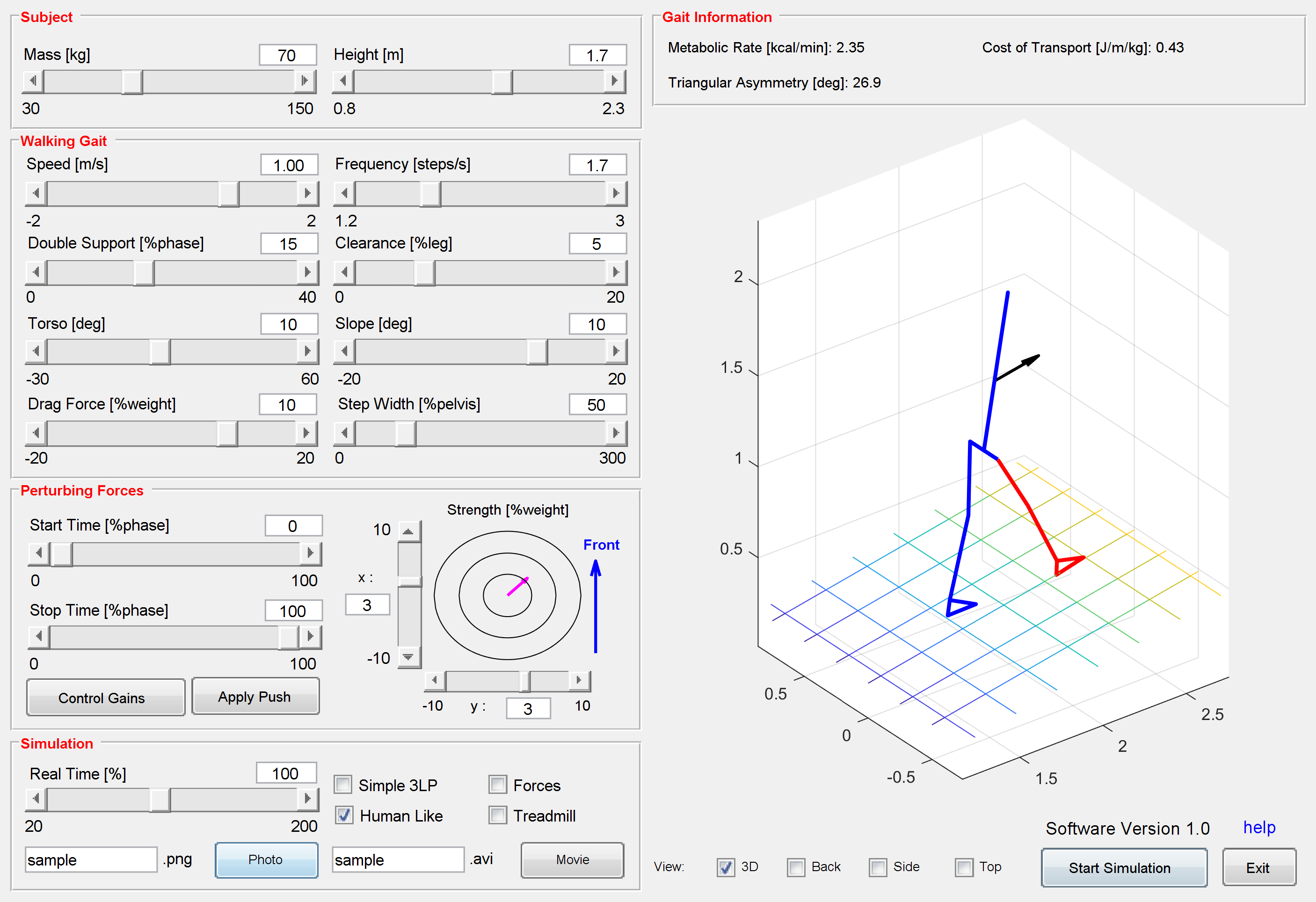}
	\caption{The GUI developed to simulate periodic and non-periodic walking conditions. This picture shows boundaries of different gait parameters within which the synthesized gait stays reasonably human-like. In this GUI, there is also a possibility to apply perturbations with different strengths and timing.}
	\label{fig::gui}
\end{figure*}

Implementation of the 3LP simulator and the kinematic conversion is currently done in MATLAB with a simple GUI (shown in Fig.\ref{fig::gui}) that allows the user to change subject/gait parameters as well as to test transient conditions \footnote{Source codes available online at \url{https://biorob.epfl.ch/research/humanoid/walkman}.}. A C++ implementation of the 3LP model and the time-projection controller is also available for robotic applications \cite{faraji20173lp3}. These codes only contain pure mathematic formulas in closed-form. The most computationally complex function in our method solves a linear system of equations to find periodic walking gaits in equations (\ref{eqn::3lp_periodic}). We use the Eigen library \cite{eigenweb} to perform this operation in microseconds. The inverse kinematic problem at the last stage of the kinematic conversion is also as simple as finding roots of a second-degree polynomial in closed-form. Using the MATLAB interface in this section, we present a collection of various walking trajectories produced with the proposed framework.

\subsection{Different Speeds}
We start this section by providing a comparison of synthesized walking trajectories against human trajectories at different walking speeds ($2,4,6\ km/h$). The data presented here is collected by a lab motion-capture system from treadmill walking. Five subjects with average height of $1.76\pm0.11 m$ and weight of $68\pm14\ kg$ participated in the experiment, walking for a minute at each desired speed to collect enough gait cycles. We measured gait parameters and replicated each experiment by our model to find corresponding synthesized gaits. Example collected and synthesized trajectories of one subject are demonstrated in Fig.\ref{fig::plot_speeds} which visually look very similar. Over all subjects and speeds, we found average correlations of $0.80\pm0.06$ for the hip angles, $0.86\pm0.04$ for the knee angles and $0.72\pm0.09$ for the ankle angles. Our method can produce many features of human walking such as pelvis vertical excursions \cite{gard2004comparison}, ground clearance \cite{wu2016determinants}, heel-toe motions \cite{cappellini2006motor} and lateral bounces \cite{donelan2002mechanical}. However, it does not produce pelvis and trunk rotations as well as foot flapping. Apart from scaling with respect to the walking speed parameter, our method supports variation of many other gait conditions as follows.

\begin{figure}[]
    \centering
    \includegraphics[trim = 0mm 0mm 0mm 0mm, clip, width=0.49\textwidth]{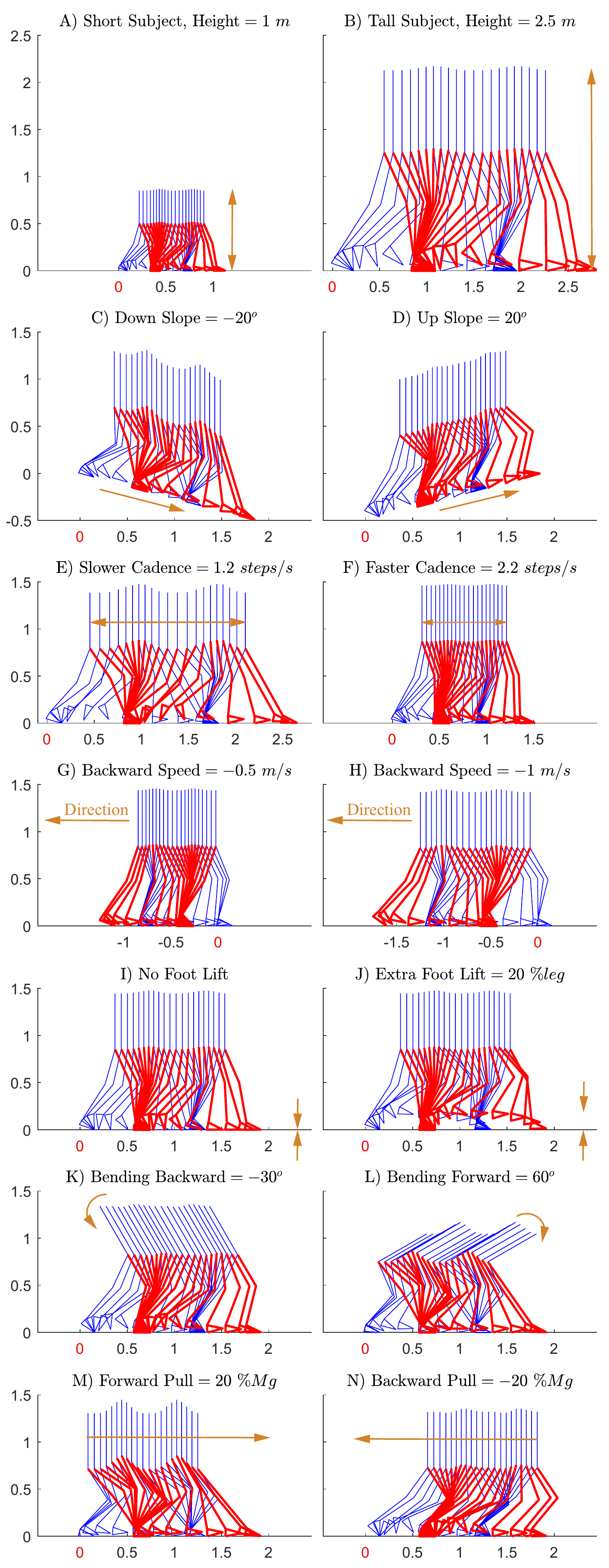}
\end{figure}
\begin{figure} []
    \caption{Gait snapshots produced at different walking conditions. By default, we simulate an adult person (height of $1.7m$, a weight of $70kg$) walking at a speed of $1m/s$, a frequency of $1.7\ steps/s$ and a ground clearance of $5\%leg$. A) and B) show walking gaits for a child and a very tall person. C) and D) demonstrate inclined walking conditions at moderate slopes. E) and F) show the effect of changing walking frequency which directly influences the step length. G) and H) show backward walking gaits. I) and J) demonstrate no foot clearance and extra foot clearance conditions. K) and L) simulate walking gaits with different torso styles. Finally, M) and N) simulate walking gaits with considerable external dragging forces. Generation of walking gaits while combining all these conditions is also possible. }
    \label{fig::plot_params}
\end{figure}

\subsection{Model Sizes} The 3LP model is scalable with respect to the mass and subject height properties. In this work, we considered average human anatomic proportions \cite{de1996adjustments} to scale all limb masses and body segments only with the overall body mass and height. However, both the 3LP model and kinematic conversion are independent of body properties without needing any re-tuning. Fig.\ref{fig::plot_params}A shows a child at the height of $1m$ performing a walking gait similar to a tall adult of $2.5m$ shown in Fig.\ref{fig::plot_params}B. In these case, we scaled the walking speed proportionally. The gait kinematics in 3LP is independent of the body mass however \cite{faraji20173lp}.

\subsection{Inclined Walking} By increasing or decreasing the terrain inclination, we can produce human-like walking gaits without re-tuning of any other parameter. The resulting kinematics shown in Fig.\ref{fig::plot_params}C,D are very similar to the human data \cite{leroux2002postural}. However, our method is not able to simulate extreme climbing cases where the hands are also involved. An interesting feature of inclined walking is extra knee flexion at the touch-down moment on positive slopes \cite{leroux2002postural} which is observed in Fig.\ref{fig::plot_params}D as well.

\subsection{Walking Frequency}
Although human walks at a particular combination of walking speeds and frequencies \cite{bertram2005constrained}, the frequency can be changed while keeping the speed constant. This directly influences the step length which is increased for example when the frequency is decreased (shown in Fig.\ref{fig::plot_params}E). While the 3LP model easily supports this modulation of frequency, our kinematic conversion method can produce realistic walking gaits in both low frequency and high frequency conditions shown in Fig.\ref{fig::plot_params}E, F. 

\subsection{Backward Walking}
The 3LP model can easily simulate backward walking by finding solutions in the linear null-space of initial gait conditions produced by equations (\ref{eqn::3lp_periodic}). The CoP motion can also be easily reverted to make the motion more realistic. This fact is reflected in the kinematic conversion method as well by automatic reversion of $p_{CoP}(t)$ and $q_{CoP}(t)$ trajectories in (\ref{eqn::pqcop}). Without changing other parameters, the 3LP model can easily walk backward at different speeds while the kinematic conversion produces human-like coordinations of lower-limb segments shown in Fig.\ref{fig::plot_params}G, H.

\subsection{Ground Clearance}
Our model simulates this motion with a simple sinusoidal curve while the actual curve in human might be slightly different, especially in extra foot lift conditions \cite{wu2016determinants}. Our simple strategy produces visually plausible walking gaits (shown in Fig.\ref{fig::plot_params}I, J) while the main inconsistency comes from missing the short flapping phase after the heel-strike. Also, an extra ground clearance might slightly affect the swing dynamics which is not included in the 3LP model.

\subsection{Torso Style}
A vast part of the walking animation literature introduces methods to produce walking gaits at different torso angles, referred to as torso styles. This is achieved via a simple proportional-derivative controller in the stance hip to regulate the torso angle while foot-placement algorithms automatically compensate the dynamic effects of such asymmetry \cite{yin2007simbicon, mordatch2010robust, coros2010generalized}. Our 3LP model can easily simulate these scenarios while the kinematic conversion adjusts the kinematics automatically, shown in Fig.\ref{fig::plot_params}K, L. Note that bending backward is uncomfortable for human while in simulations, it is theoretically possible. The extra vertical excursion observed in Fig.\ref{fig::plot_params}L is also less human-like. When bending forward, humans damp these vertical excursions by an increased flexion in the knees at the mid-stance moment \cite{grasso2000interactions}, probably for the sake of comfort or gaze stabilization. Our method, however, produces a peak in the pelvis height trajectory at this moment which results in a stretched-knee posture. This probably prevents our method to simulate extreme torso bending conditions. 

\begin{figure*}[]
    \centering
    \includegraphics[trim = 0mm 0mm 0mm 0mm, clip, width=1\textwidth]{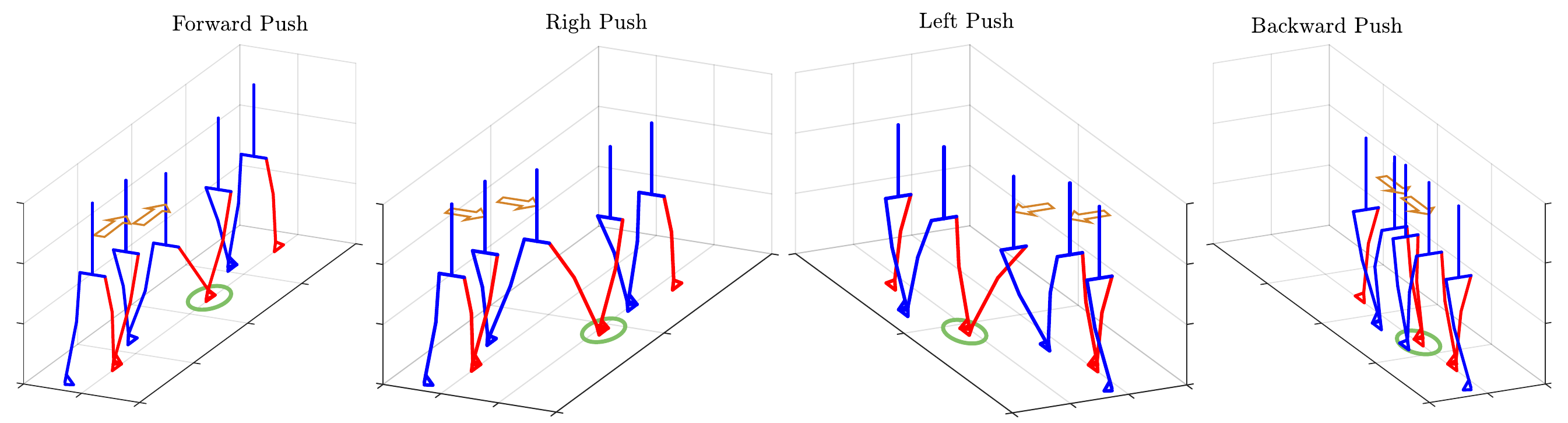}
    \caption{Different scenarios of perturbed walking conditions where the robot is subject to large external pushes of $50N$ applied continuously during a step phase. Our time-projecting controller can easily stabilize the 3LP model while the kinematic conversion takes the 3LP state and produces human-like trajectories. Pure interpolation methods for walking animation cannot simulate perturbed walking conditions interactively while physics-based animations require a lot of computation power to simulate such interactions with the environment. Our hybrid approach, however, can cover a wide range of transient walking conditions.}
    \label{fig::plot_pushes}
\end{figure*}

\subsection{Dragging Forces}
Another interesting scenario is to produce periodic walking gaits subject to constant external dragging forces. This could be useful in a simulation of pulling or pushing heavy objects \cite{coros2010generalized}. 3LP can easily produce such walking gaits by including the external force in symbolic equations. We considered forces applied to the torso while 3LP formulations can be easily changed to simulate other force application points. Although the triangular coordination between the two legs in 3LP becomes asymmetric in these conditions, the kinematic conversion can still produce lower-limb coordinations adaptively. 

\subsection{Push Recovery}
The main purpose of incorporating a physics-based animation in our method is to model interactions with the environment. In addition to simulating constant external forces discussed previously, we are interested in simulating transient conditions due to disturbances as well. This simulation scenario involves time-integration, i.e., considering small time-steps, applying arbitrary disturbance forces at each time-step and finding system evolution through integration. Thanks to the linear equations of the 3LP model, the system evolution can be described by a single closed-form matrix which relieves the need to perform iterations. Also, if the disturbance pattern is known beforehand, we can find closed-form equations and avoid using small time-steps, depending on the precision required. Fig.\ref{fig::plot_pushes} demonstrates transient walking conditions due to external pushes applied in different directions. While the 3LP model and the time-projecting controller can produce natural and stable horizontal motions, the kinematic conversion takes the 3LP state and produces vertical motions adaptively.

The strength of our method lies in generating walking trajectories in different combinations of all the previously-mentioned gait conditions. We limit our results section to discuss each gait condition separately. However, thanks to the closed-form solutions available, changing many gait conditions at the same time does not need any re-tuning of trajectory generation or control parameters. The next section will provide a comprehensive discussion of these strengths and intrinsic limitations of the proposed approach.

\section{Discussion}
The proposed method combines physics-based and pure interpolation approaches in the literature for walking trajectory generation. We simulate physics by a linear simplified model called 3LP that has closed-form solutions. On top of this model, in the present work, we propose an adaptive kinematic converter which synthesizes human-like lower-limb postures. The resulting trajectories follow the overall dynamics of 3LP while remaining geometrically feasible in transient conditions. The goal of such a hybrid approach is to achieve faster simulation speeds while offering an online walking control. We can simulate interactions with the environment to some extent and produce transient walking trajectories thanks to a previously developed walking controller called time-projection. This controller together with the 3LP model encapsulates important dynamic properties and control rules needed to stabilize the gait in a wide range of walking conditions. Therefore, the proposed model-based approach does not have any parameter to tune. 

\subsection{Closed-Form Solutions}
The 3LP model \cite{faraji20173lp} and the time-projection controller \cite{faraji20173lp2} were originally developed to extend the LIP model and MPC control paradigm \cite{faraji2014robust} for humanoid walking application \cite{faraji20173lp3}. The closed-form equations of 3LP or LIP enable MPC controllers to stabilize the system in an online fashion by adjusting footstep locations. In a previous work based on the LIP model, our MPC controller was able to solve a quadratic optimization problem in less than a millisecond and suggest footstep corrections online \cite{faraji2014robust}. The time-projection control, however, aims at finding closed-form solutions for the numeric optimizations of MPC as well. Therefore, the combination of a linear model and time-projection control can offer simulation speeds as fast as microseconds. Besides, thanks to all these closed-form solutions, we do not need to use sub-millisecond simulation time-steps like \cite{yin2007simbicon, mordatch2010robust, coros2010generalized} to ensure numerical stability. To generate animations, we only need to consider movie frames (e.g., $30$ frames per second) and find the system evolution in between by closed-form matrices. 

While walking gait generation and stabilization with a simplified model are developed in our previous work, in the present paper, we aimed at filling the gap with reality. In other words, we proposed a kinematic conversion method to convert walking trajectories from the 3LP space to a real character with thigh, shank and foot segments. The novelty of this paper, therefore, lies in the conversion method and the entire architecture that produces periodic and transient human-like walking trajectories. Using an intuitive mixture of the geometric variables in 3LP, we can produce smooth vertical excursions and human-like thigh-shank-foot coordinations. Although we do not simulate dynamics of these leg segments explicitly, each leg follows the approximate dynamics encoded in the 3LP model. Following the same philosophy of developing closed-form solutions in the 3LP model and the time-projection controller, the kinematic conversion is also formulated in closed-form. Various parametric trajectory design or interpolation methods already exist in the literature and offer a similarly fast simulation speed, but they cannot produce transient walking conditions. They also need either a large library of human trajectories to interpolate or a large set of trajectory or control parameters to produce as many walking conditions. The proposed architecture is mathematically involved, but generic and straightforward to be used in walking control, animation or analysis. 

\subsection{Limitations}
Walking dynamics in our method is simulated in the 3LP model which relies on linearization assumptions. 3LP is an extension of the LIP model and both assume linear pendular dynamics. 3LP can simulate swing and torso balancing dynamics (in addition to falling dynamics of the LIP model) which allow for simulation of faster walking gaits. However, the resulting motions are valid only where the coupling between horizontal and vertical dynamics is negligible. Our method cannot simulate very large step lengths. Besides, the time-projection controller does not consider such feasibility boundaries. Although this controller always stabilizes the gait, in extreme conditions, it might produce large step lengths that violate decoupling assumptions. These conditions only happen in case of very large disturbances or considerable sudden changes in the desired gait parameters (such as speed or frequency). The linearity assumptions allow for simulation of flat or inclined walking conditions, but not uneven terrains or structured environments. However, if the terrain profile is known in advance, we can design certain height change profiles and solve the new linear time-variant 3LP equations numerically. Besides, the 3LP model does not simulate turning in the current implementation due to nonlinearities. We can remove the pelvis width and allow the 3LP model to turn, but an artificial separation of the two feet is needed \cite{faraji2014robust}. We consider these fundamental improvements for future work.

The present framework can simulate different gait conditions as shown in Fig.\ref{fig::plot_params}, however, some transient conditions are not always easy to model. We cannot simulate torso oscillations unless we linearize the torso and give it a degree of freedom. In this work, the 3LP model assumes a fixed torso angle and finds necessary stance hip torques to realize this assumption. Variations in the external dragging forces are easy to simulate though, since they can be treated as perturbations. A transient change of speed, step frequency and double support time is possible in our current framework. The time-projecting controller can handle them stably. 

While the entire framework can simulate a wide and continuous range of walking conditions with different combinations, the natural human-like choice of walking parameters remains un-modeled. Humans can also walk at various gait conditions, but not necessarily be energy optimal or comfortable. When changing the walking speed, humans change the frequency \cite{bertram2005constrained}, double support ratio \cite{cappellini2006motor}, ground clearance \cite{ivanenko2002control} and torso orientation \cite{song2012regulating}. All these parameters change in inclined walking as well \cite{vogt1999measurement}. Our framework provides the necessary platform to simulate all these walking conditions, but not including human-optimal relations of parameters. Realistic choices of these parameters can be extracted from the related biomechanics literature in walking animations like \cite{boulic1990global, li2000simulating}. Other human-like walking features such as arm motions or pelvis rotations can be added easily without affecting the overall walking dynamics \cite{boulic1990global}.

\subsection{Applications and Future Work}
In a trade-off with some of the features offered by physics-based frameworks like rough-terrain locomotion \cite{mordatch2010robust, coros2010generalized}, we achieved much faster simulation speeds by simplifying the physical model. Our method provides pure mathematical formulas with a minimal dependency on the Eigen library for a matrix inversion. Our source codes can be easily integrated with other simulators to produce animations on visually more human-like characters. It can be used for crowd-walking simulations as well as animations on portable electronic devices with a limited computational power. Besides, the ideas introduced in this paper can be used to control humanoids or simulated robots. In particular, our kinematic conversion can be used to produce more human-like pelvis trajectories and thigh-shank-foot coordinations. All the source codes would be freely available online after publication. 

\begin{acks}
    This work was funded by the WALK-MAN project (European Community's 7th Framework Programme: FP7-ICT 611832).
\end{acks}

\bibliographystyle{ACM-Reference-Format}
\bibliography{Biblio}

\end{document}